\documentclass[letterpaper, 10 pt, conference]{ieeeconf}  

\IEEEoverridecommandlockouts                              

\usepackage{amsmath,amsfonts}
\usepackage{algorithmic}
\usepackage{array}
\usepackage[caption=false,font=normalsize,labelfont=sf,textfont=sf]{subfig}
\usepackage{textcomp}
\usepackage{stfloats}
\usepackage{url}
\usepackage{verbatim}
\usepackage{graphicx}
\def\BibTeX{{\rm B\kern-.05em{\sc i\kern-.025em b}\kern-.08em
    T\kern-.1667em\lower.7ex\hbox{E}\kern-.125emX}}
\usepackage{balance}
\usepackage[pagebackref,breaklinks,colorlinks,allcolors=cvprblue]{hyperref}
\usepackage{multirow}
\usepackage[dvipsnames]{xcolor}
\hypersetup{
    colorlinks=true,
    linkcolor=red,
    filecolor=green,      
    urlcolor=magenta,
    citecolor=green,
}

\usepackage{algorithm}
\usepackage{algorithmic}
\usepackage{booktabs}
\title{\LARGE \bf
AstroRAG - A Pagerank-Based Retrieval-Augmented Generation Pipeline for Question Answering in Astronomy
}

\author{Zhifeng Wang$^{1,2}$ and Jason Jingshi Li$^{2}$ and Kaihao Zhang$^{3}$ and Ramesh Sankaranarayana$^{1}$  
\thanks{*This work was supported by Learning Machines Pty Ltd during the internship*}
\thanks{$^{1}$Zhifeng Wang and Ramesh Sankaranarayana are with
        Australian National University, Canberra, ACT, Australia
        {\tt\small zhifeng.wang,ramesh.sankaranarayana@anu.edu.au}}%
\thanks{$^{2}$Jason Jingshi Li is with Learning Machines Pty Ltd, Canberra, ACT, Australia
        {\tt\small jason@learningmachines.au}}%
\thanks{$^{3}$Kaihao Zhang is with
        Australian National University, Canberra, ACT, Australia
        {\tt\small super.khzhang@gmail.com}}%
}

\begin{document}

\maketitle
\thispagestyle{empty}
\pagestyle{empty}

\begin{abstract}
Large language models (LLMs) demonstrate strong performance in natural language processing but often generate factual errors when relying solely on parametric knowledge. Retrieval-Augmented Generation (RAG) mitigates these errors by grounding responses in external evidence, yet conventional retrieve-and-dump approaches frequently introduce irrelevant context that degrades answer quality. In this work, we present AstroRAG - a PageRank-based retrieval-augmented generation (RAG) pipeline adapted for question answering in astronomy. The system performs token-aware chunking and per-instance, ephemeral indexing in Elasticsearch, then executes a two-stage retrieval: (i) Maximal Marginal Relevance (MMR) to obtain a small, diverse candidate set and (ii) a reader-driven PageRank (PR) re-ranking on a similarity graph to identify a compact, mutually supportive context under a strict token budget. Our design is training-free, privacy-preserving, and reproducible, as each instance is processed through transient indexing to prevent cross-task leakage. We evaluate the pipeline on the AstroQA benchmark for astronomy QA, and demonstrate competitive performance across all difficulty levels. In particular, the RAG-enhanced Mistral-7B achieves \textbf{79.49\% accuracy} and \textbf{79.49\% F1-score}, nearly doubling the performance of its non-RAG counterpart. These results highlight the effectiveness of disciplined retrieval and refinement in boosting domain-specific reasoning, establishing a robust foundation for extending RAG to other scientific fields.
\end{abstract}

\section{Introduction}
Large language models (LLMs) \cite{bai2025qwen2, touvron2023llama, nguyen2023astrollama} demonstrate strong performance across diverse natural language tasks, yet they remain vulnerable to factual errors when relying solely on parametric knowledge. Retrieval-Augmented Generation (RAG) mitigates these errors by conditioning generation on external evidence \cite{de2023fido}. In practice, however, many ``retrieve-and-dump'' pipelines fetch a fixed number of passages irrespective of necessity, introducing distractors that degrade answer quality—particularly in domain settings such as astronomy \cite{xu2025evaluating}—and sometimes producing responses that are not fully consistent with the cited sources.

Recent studies \cite{ni2025towards, abras2025can} clarify these limitations and suggest design principles for more reliable systems. Controlled long-context evaluations reveal a robust U-shaped positional effect: models are most accurate when key evidence appears at the beginning or the end of the prompt and perform substantially worse when it is placed in the middle, a pattern that persists even for long-context variants \cite{liu2024lost}. Systems such as M-RAG \cite{wang2024m} further reduce interference by reasoning over documents separately—producing per-document partial answers—and subsequently consolidating them. These findings indicate that effective RAG should regulate not only how much evidence is retrieved, but also where it is positioned and how it is consumed. In this context, astronomical question answering presents unique retrieval and reasoning challenges due to the field's dense technical terminology, extensive use of mathematical formalism, and reliance on precise contextual grounding. Prior work such as Pathfinder \cite{iyer2024pathfinder} demonstrates the potential of retrieval-augmented systems in exploring the astronomy literature while also revealing persistent limitations—particularly in handling highly specialized or equation-rich content, the gap between general-domain and astronomy-specific vocabulary, and the need for interpretability and traceability in scientific reasoning. These challenges motivate domain-aware adaptations of RAG pipelines for astronomy, emphasizing verifiability, privacy, and reproducibility.

\begin{figure*}[htbp]
  \centering
  \includegraphics[width = \linewidth]{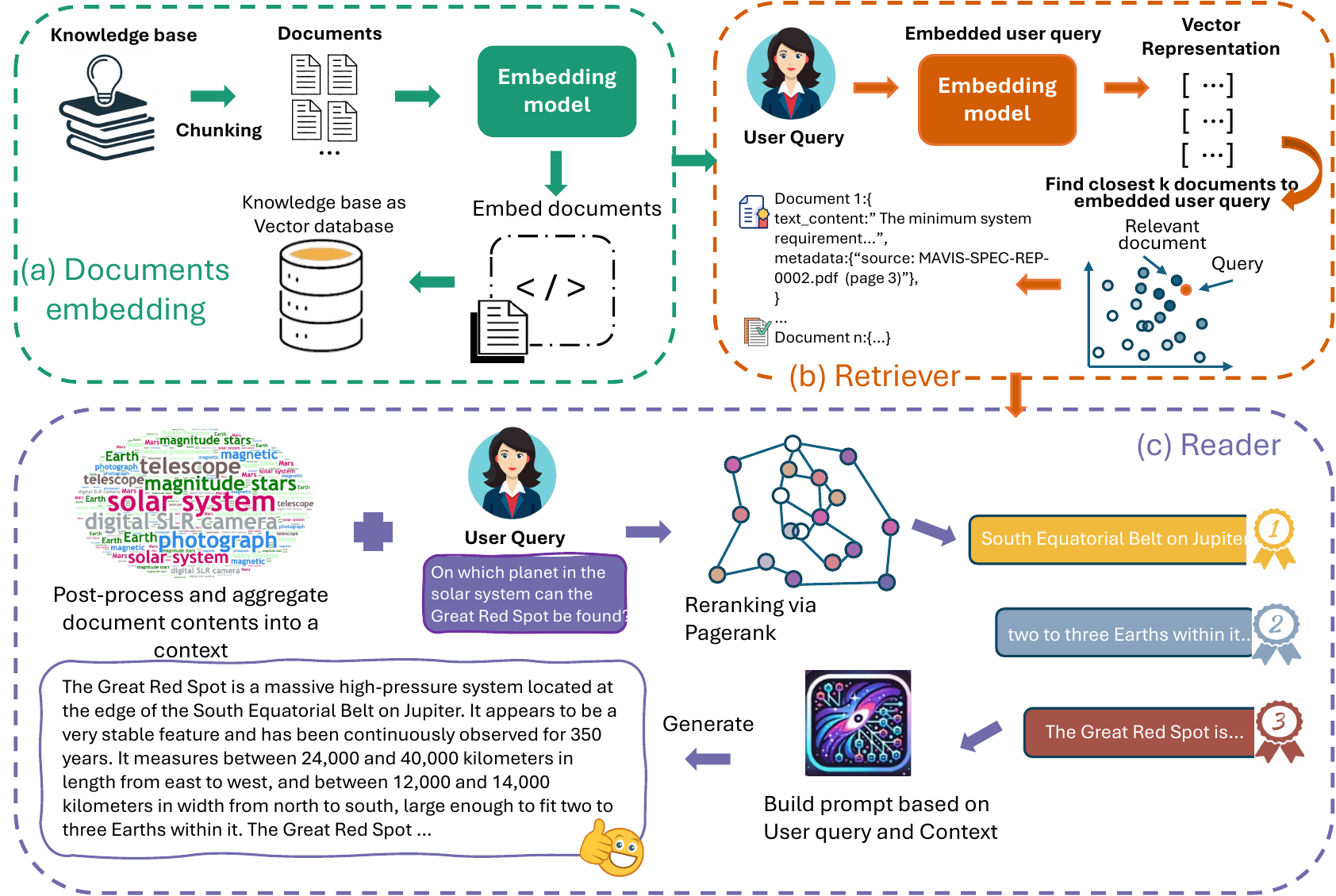}
  \vspace{-1em}
  \caption{\textbf{Pagerank-based RAG workflow for astronomy QA.} Domain documents are chunked and embedded, then indexed in a vector knowledge base. At query time, the retriever embeds the user question and performs similarity search to return the top-k relevant passages with source metadata. The reader re-rank via Pagerank and aggregates these passages into context and builds a prompt for the LLM, which generates a grounded answer (with traceable sources).}
  \label{rag_workflow}
\end{figure*}

To address these challenges, we introduce AstroRAG, a lightweight, fully local RAG framework tailored to question answering in astronomy. As illustrated in Fig.~\ref{rag_workflow}, the system performs token-aware chunking of the input corpus, embeds these chunks, and indexes them in a transient (per-instance) Elasticsearch store. At query time, we execute a two-stage retrieval: Maximal Marginal Relevance (MMR) to obtain a small, diverse candidate set, followed by a reader-driven \emph{PageRank} re-ranking on a similarity graph to identify a compact, mutually reinforcing context. The final set of PR-selected snippets is concatenated into a single prompt, and a local LLM produces a concise answer along with the exact supporting source passages.

The contributions are listed below: (1) Instance-level isolation and traceability: we ingest only the relevant file(s), construct a transient Elasticsearch index per instance, record the answer and its supporting snippets, and erase the index post-answer to avoid contamination. (2) Two-stage retrieval with reader-driven PR: we pair MMR with PageRank re-ranking to select a small, coherent set of snippets (default small $k$) that balances query relevance and inter-snippet support, then trim to a fixed token budget for faithful, compact prompts. (3) Fully local, modular pipeline: embeddings, retrieval, and generation run on local hardware, support optional HyDE, and interface with multiple local LLM backends, enabling offline operation and privacy-sensitive use.

\section{Related Work}
\subsection{Long-Context Large Language Models}
Large Language Models (LLMs) show strong performance across complex reasoning tasks, including specialized domains like astronomy. Advances in long-context modeling—such as FlashAttention \cite{dao2022flashattention}, Longformer \cite{beltagy2020longformer}, and Mistral \cite{samo2024fine}—extend context windows with improved efficiency. However, recent evaluations reveal persistent limitations: Liu \textit{et al.} \cite{liu2024lost} identify a U-shaped positional effect, where models perform best when evidence appears at the beginning or end of the prompt, with reduced accuracy in the middle. In retrieval-augmented generation (RAG), simply retrieving more passages often saturates reader performance, suggesting that reranking and truncation are more effective than naive context expansion. Domain-specific models like AstroLLaMA \cite{nguyen2023astrollama} improve scientific grounding through targeted fine-tuning but remain constrained by base model context limits.

\subsection{Retrieval-augmented Generation models}
Retrieval-Augmented Generation (RAG) enhances LLMs by grounding responses in external evidence and has evolved across three main paradigms: naïve, advanced, and modular. Early naïve systems follow a simple retrieve–read approach—indexing text, retrieving top-k chunks by similarity, and inserting them into the prompt—but suffer from retrieval errors, context overload, and hallucinations \cite{chen2017reading}. Advanced RAG systems address these issues through query reformulation, improved indexing, reranking, and context compression. For instance, Ma \textit{et al.} \cite{ma2023query} use RL-based query rewriting to improve evidence retrieval and downstream QA performance, while BEQUE \cite{peng2024large} enhances e-commerce search using multi-stage query rewriting and feedback alignment. Modular RAG systems further extend flexibility by introducing pluggable components—like explicit search, routing, or memory—and supporting iterative or end-to-end workflows. KnowledGPT \cite{wang2023knowledgpt} enables structured KB querying through program-of-thought prompting, and RAG-Fusion \cite{rackauckas2024rag} improves retrieval via multi-query fusion and reranking, leading to more accurate, multi-perspective answers.

\section{Methods}
We propose a lightweight, fully local Retrieval-Augmented Generation (RAG) pipeline tailored for domain-specific question answering over local \texttt{.txt}, \texttt{.pdf}, and \texttt{.md} sources. The system integrates token-aware chunking, semantic retrieval, \emph{PageRank} (PR) re-ranking, and local large language models (LLMs) to produce concise, source-grounded responses. We implement the pipeline as a Streamlit application to enable simple deployment and interactive experimentation.
\subsection{Pipeline Overview}
Our pipeline follows a per-instance strategy. For each question, we (i) load and preprocess the associated background documents, (ii) ingest the resulting chunks into a transient Elasticsearch index, (iii) retrieve a small candidate set using Maximal Marginal Relevance (MMR), (iv) apply a reader-driven PageRank re-ranking to select the final set of supporting passages, (v) generate the answer using a local LLM, and (vi) reset the index (delete-by-query) to prevent cross-instance leakage. For traceability, we record the model's answer, the supporting source document filenames, and the retrieved text snippets.

\subsection{Document Preprocessing and Chunking}
Each document $T_i$ is normalized to remove layout artifacts (e.g., converting single line breaks to spaces). We then split it with a recursive, token-aware text splitter using a \texttt{tiktoken}-based length function $\ell(\cdot)$ with encoding \texttt{p50k\_base}, maximum chunk size $L{=}300$ tokens and an overlap of $\Delta{=}30$ tokens:
\begin{equation}
S(T_i) = \{c_{i1}, \dots, c_{im}\},
\label{document_preprocessing}
\end{equation}
\[
\text{such that}\quad \ell(c_{ij}) \le L,\quad \text{overlap}(c_{ij}, c_{i,j+1})=\Delta.
\]
Each chunk carries metadata (filename, page, chunk ID) for provenance and traceability. To enable page-level metadata, PDFs are parsed page-wise with \texttt{pypdf} before chunking.

\subsection{Embedding and Vector Indexing}
We embed chunks with a sentence embedding model $f_\theta$  such as Sentence-BERT \cite{reimers-2019-sentence-bert}. For each chunk $c_{ij}$:
\[
e_{ij} = f_\theta(c_{ij}) \in \mathbb{R}^d.
\]
Optionally, using HyDE, a local LLM $g_\phi$ generates a hypothetical passage $\tilde{t}_i=g_\phi(q_i)$ which we embed as $\tilde{e}_i=f_\theta(\tilde{t}_i)$ and use as the query embedding to improve retrieval on sparse queries. All chunk embeddings for the current instance are stored in the ephemeral Elasticsearch index via a LangChain.

\subsection{Query Construction, Two-Stage Retrieval, and PageRank Re-Ranking}
For multiple-choice QA, a unified prompt string is formed by concatenating the question $q_i$ with the options $O_i=\{A_i,B_i,C_i,D_i\}$ and a brief instruction $I$:
\[
p_i=\operatorname{concat}\!\Big(q_i,\ \operatorname{format}(O_i),\ I\Big).
\]
For retrieval, we compute the query embedding $e_i^{(q)}$ either directly from $q_i$ (default) or from HyDE's \cite{gao2023precise} $\tilde{t}_i$.

\paragraph{Stage 1: MMR retrieval.}
We use cosine similarity $s(u, v) = \frac{u^\top v}{\|u\| \|v\|}$ to measure relevance. Using Maximal Marginal Relevance (MMR) with $k=3$, we select a diverse, relevant candidate set. The standard MMR objective for selecting the next document $d^*$ from candidates $D\!\setminus\! S$ is
\[
d^*=\arg\max_{d\in D\setminus S}\Big[\lambda\,s(d,q) - (1-\lambda)\max_{d'\in S}s(d,d')\Big],
\]
with $\lambda \in [0,1]$ controlling the relevance–diversity trade-off. This selection is repeated until $k$ items are chosen.

\paragraph{Stage 2: Reader re-ranking via PageRank.}
Let the $k$ retrieved chunks have embeddings $\{x_1,\dots,x_k\}$. We build a similarity graph using cosine similarity $S_{ij}=s(x_i,x_j)$, setting $S_{ii} = 0$ and thresholding edges below $\tau=\texttt{pr\_min\_sim}$ (default $\tau{=}0.01$):
\[
W_{ij}=\begin{cases}
S_{ij} & \text{if } S_{ij}\ge \tau\\
0 & \text{otherwise.}
\end{cases}
\]
Rows are normalized to form a transition matrix $P$; if a row sum is zero, we use a uniform row:
\[
P_{ij}=
\begin{cases}
\dfrac{W_{ij}}{\sum_k W_{ik}} & \text{if } \sum_k W_{ik}>0\\[6pt]
\dfrac{1}{k} & \text{otherwise.}
\end{cases}
\]
The personalization vector $v$ is derived from non-negative query--chunk cosine scores $s_i=\max\!\big(0,\,s(x_i,e_i^{(q)})\big)$, normalized to sum to one. We iterate PageRank with damping $\alpha={pr_\alpha}$ (default $\alpha{=}0.85$):
\[
r^{(t+1)}=\alpha P^\top r^{(t)} + (1-\alpha)v,\qquad r^{(0)}=\tfrac{1}{k}\mathbf{1}.
\]
We select the top $\texttt{pr\_top\_k}$ (default $=3$) chunks by $r$ to form the final context. To fit the LLM context window, we then trim the concatenated chunks to a token budget $T{=}1800$ using the \texttt{tiktoken} length $\ell(\cdot)$.

\subsection{LLM-Based Answer Generation}
 We support multiple local LLMs via different backends such as AstroSage Chat (8B) — accessed via OpenAI-compatible API, Mistral Chat (7B), LLaMA 2 (13B) — served through HuggingFace Text Generation Inference (TGI).

Given the final PR-selected context $D^*=\{d_1,\dots,d_{k^*}\}$ (ordered by $r$), we construct a single compact prompt:
\[
\texttt{Context:\ {[Chunk 1]$...$[Chunk $k^\star$]}\ Question:\ }p_i,
\]
and generate the answer in one pass:
\[
\hat{A}=\arg\max_{a\in\mathcal{V}^*} p_\phi\big(a \mid p_i,\ D^\star\big).
\]
Decoding follows the code defaults:
\[
p_\phi(w_t\mid\cdot)=\operatorname{softmax}\!\Big(\tfrac{z_t}{\tau_{\text{dec}}}\Big),\quad
\tau_{\text{dec}}=0.005,\ \ \texttt{top\_p}=0.95,
\]
with \texttt{max\_new\_tokens}$=128$ for all backends. The final output is $\hat{A}$ along with the filenames and chunk texts used.
\begin{table*}[htb]
\centering
\small
\caption{\textbf{Comparison of zero-shot question-answer performance across various LLMs on the AstroQA dataset \cite{li2025astronomical}}, including Llama 2-Chat \cite{touvron2023llama}, Mistral-7B \cite{samo2024fine}, AstroSage-Llama-3.1-8B \cite{de2025achieving}, and the proposed RAG-enhanced Llama 2-Chat model, Mistral-7B  and AstroSage-Llama-3.1-8B model. Results are reported for accuracy (Acc\%) and F1-Score (F@1) across datasets with Easy, Medium, and Hard difficulty levels.}
\resizebox{\linewidth}{!}{%
\begin{tabular}{c c c c c c c c c c}
\toprule
\multirow{2}{*}{\textbf{Methods}} & \multirow{2}{*}{\textbf{Backbone}} & \multicolumn{2}{c}{\textbf{Easy}} & \multicolumn{2}{c}{\textbf{Medium}} & \multicolumn{2}{c}{\textbf{Hard}} & \multicolumn{2}{c}{\textbf{Total}} \\ 
\cmidrule(lr){3-10}
& & \textbf{Acc(\%)} & \textbf{F@1} & \textbf{Acc(\%)} & \textbf{F@1}  & \textbf{Acc(\%)} & \textbf{F@1}  & \textbf{Acc(\%)} & \textbf{F@1}  \\ 
\midrule
\textbf{Llama 2-Chat \cite{touvron2023llama}} & MPT & 28.24 & 24.06 & 28.45 & 22.36 & 27.10 & 21.01 & 28.06 & 23.10 \\
\textbf{Mistral-7B \cite{samo2024fine}} & ViT & 20.74 & 8.60 & 21.97 & 9.57 & 24.05 & 10.26 & 21.74 & 9.20 \\
\textbf{AstroSage-Llama-3.1-8B \cite{de2025achieving}} & MPT & 41.47 & 41.08 & 43.94 & 43.92 & 39.31 & 38.53 & 41.71 & 41.45 \\
\midrule
\textbf{Llama 2-Chat \cite{touvron2023llama}+AstroRAG (Ours)}  & MPT & 52.94 & 53.63 & 49.30 & 49.72 & 54.20 & 54.07 & 52.20 & 52.78\\
\textbf{AstroSage-Llama-3.1-8B \cite{de2025achieving}+AstroRAG (Ours)}  & MPT & \textbf{\textcolor{blue}{76.03}} & \textbf{\textcolor{blue}{76.15}} & \textbf{\textcolor{blue}{77.46}} & \textbf{\textcolor{blue}{77.47}} & \textbf{\textcolor{blue}{81.30}} & \textbf{\textcolor{blue}{81.20}} & \textbf{\textcolor{blue}{77.49}} & \textbf{\textcolor{blue}{77.59}} \\
\textbf{Mistral-7B \cite{samo2024fine}+AstroRAG (Ours)}  & ViT & \textbf{\textcolor{red}{77.50}} & \textbf{\textcolor{red}{77.45}} & \textbf{\textcolor{red}{80.28}} & \textbf{\textcolor{red}{80.27}} & \textbf{\textcolor{red}{83.59}} & \textbf{\textcolor{red}{83.50}} & \textbf{\textcolor{red}{79.49}} & \textbf{\textcolor{red}{79.49}} \\
\bottomrule
\end{tabular}
}
\label{model_performance_table}
\end{table*}
\subsection{Batch Evaluation Protocol}
To evaluate multiple-choice question answering (QA) at scale, we adopt the following batch evaluation protocol. For each QA instance, we first normalize the text and split it into small, overlapping chunks using a \texttt{tiktoken}-based length function. Metadata is attached to each chunk, and the collection is ingested into an ephemeral Elasticsearch index to enable efficient retrieval. Next, we construct a query string by concatenating the question, its answer choices, and a concise instruction. A small candidate set is retrieved using Maximal Marginal Relevance (MMR), which balances relevance and diversity. We then build a similarity graph where each retrieved chunk becomes a node and edges connect nodes with semantic similarity above threshold $\tau$. To further refine the candidate set, we apply a reader-driven PageRank algorithm. This involves a random walk that spreads credit along strong semantic connections while also favoring nodes close to the query. The balance between exploration and bias is governed by the parameter \texttt{$pr_{\alpha}$}. The process runs iteratively until the scores stabilize, yielding a ranked list of chunks that are both individually relevant and mutually supportive. We then select the top $k$ chunks to form a compact and coherent context, trimming the combined content to fit within a fixed token budget. Using this context, the model generates a prediction $\hat{a}_i$. The predicted answer, along with its supporting evidence (retrieved chunks and metadata), is recorded for downstream analysis. To prevent contamination between QA instances, the temporary Elasticsearch index is deleted after each item is processed. The final output includes the model's prediction, supporting context, and associated metadata, enabling traceable and reproducible evaluation across different models and QA tasks.

\begin{table*}[tbp]
\centering
\small
\caption{\textbf{Performance comparison of RAG-based question answering on the AstroQA dataset \cite{li2025astronomical}}. We evaluate multiple models including the baseline Mistral-7B \cite{samo2024fine}, DrQA \cite{liu2024lost}, Map-Reduce \cite{de2023fido}, and our proposed RAG-enhanced variants: Mistral-7B + AstroRAG. Results are reported in terms of accuracy (Acc\%) and F1-score (F\@1) across question subsets categorized by difficulty: Easy, Medium, and Hard.}
\resizebox{\linewidth}{!}{%
\begin{tabular}{c c c c c c c c c }
\toprule
\multirow{2}{*}{\textbf{Methods}} & \multicolumn{2}{c}{\textbf{Easy}} & \multicolumn{2}{c}{\textbf{Medium}} & \multicolumn{2}{c}{\textbf{Hard}} & \multicolumn{2}{c}{\textbf{Total}} \\ 
\cmidrule(lr){2-9}
& \textbf{Acc(\%)} & \textbf{F@1} & \textbf{Acc(\%)} & \textbf{F@1}  & \textbf{Acc(\%)} & \textbf{F@1}  & \textbf{Acc(\%)} & \textbf{F@1}  \\ 
\midrule
\textbf{Baseline \cite{de2025achieving}} & 20.74 & 8.60 & 21.97 & 9.57 & 24.05 & 10.26 & 21.74 & 9.20 \\
\textbf{DrQA \cite{liu2024lost}} & 62.06 & 62.43 & 58.87 & 58.87 & \textbf{\textcolor{blue}{62.60}} & \textbf{\textcolor{blue}{62.37}} & 61.30 & 61.55 \\
\textbf{Map-Reduce \cite{de2023fido}} & \textbf{\textcolor{blue}{64.56}} & \textbf{\textcolor{blue}{64.94}} & \textbf{\textcolor{blue}{61.97}} & \textbf{\textcolor{blue}{62.39}} & 60.69 & 60.98 & \textbf{\textcolor{blue}{63.07}} & \textbf{\textcolor{blue}{63.48}} \\
\textbf{AstroRAG (Ours)}  & \textbf{\textcolor{red}{77.50}} & \textbf{\textcolor{red}{77.45}} & \textbf{\textcolor{red}{80.28}} & \textbf{\textcolor{red}{80.27}} & \textbf{\textcolor{red}{83.59}} & \textbf{\textcolor{red}{83.50}} & \textbf{\textcolor{red}{79.49}} & \textbf{\textcolor{red}{79.49}} \\
\bottomrule
\end{tabular}
}
\label{model_performance_table_sota}
\end{table*}
\section{Experiments}
\subsection{Models and Settings}
Our approach does not require training any models. We evaluate the proposed method's performance in a zero-shot setting using LLMs.  Specifically, we evaluate the capabilities of several advanced LLMs, including commercial models such as Llama 2-Chat~\cite{touvron2023llama}\footnote{https://huggingface.co/meta-llama/Llama-2-13b} and open-sourced models including Mistral-7B \cite{samo2024fine}\footnote{https://huggingface.co/mistralai/Mistral-7B-v0.1} and AstroSage-Llama-3.1-8B \cite{de2025achieving}\footnote{https://huggingface.co/AstroMLab/AstroSage-8B}. 
\subsection{Dataset details}

The AstroQA dataset \cite{li2025astronomical} is the first benchmark specifically designed for evaluating large language models (LLMs) in the field of astronomy. It contains 3,082 questions in both English and Chinese, covering six question types with corresponding gold-standard answers and supporting material. The questions span several key branches of astronomy, including astrophysics, astrometry, celestial mechanics, history of astronomy, and astronomical techniques and methods, thereby providing comprehensive domain coverage. To enable fair evaluation across both objective and subjective questions, the authors propose a novel metric called DGscore, which combines multiple evaluation measures and applies type- and difficulty-specific weighting coefficients. Validated on open-source and commercial LLMs, AstroQA is a benchmark for evaluating instruction-following, domain reasoning, and language generation in astronomy, providing a solid basis for calibrating and advancing astronomy-focused LLMs.

 Unlike existing benchmarks that rely primarily on exam questions, our AstroRef-QA is a high-quality QA dataset created through a reproducible pipeline. We chunk PDF documents into token-bounded segments, extract up to three salient topics per chunk using an LLM, and generate QA pairs guided by Bloom's Taxonomy. Near-duplicates are removed via bigram-overlap filtering. Each QA pair is evaluated on four dimensions (Relevance, Quality, Depth, Accuracy) using LLM-as-judge scoring and embedding-based similarity to its source chunk, yielding a combined quality score. Curation ensures cognitive diversity by maintaining minimum representation across Bloom levels, then admitting high-scoring pairs until reaching the target size. Every item includes provenance metadata (source file and chunk IDs) and evaluation summaries for transparency. The pipeline is configurable, resumable, and supports optional human review via a Streamlit interface, with output in JSON and CSV formats.

\begin{table}[tb]
\centering
\small
\caption{\textbf{Comparison of zero-shot question-answer performance across various LLMs on our AstroRef-QA dataset}, including Mistral-7B \cite{samo2024fine}, AstroSage-Llama-3.1-8B \cite{de2025achieving}, and the proposed RAG-enhanced Mistral-7B  and AstroSage-Llama-3.1-8B model. Results are reported for ROUGE-L and ROUGE-N.}
\resizebox{\linewidth}{!}{%
\begin{tabular}{c c c c c c c c}
\toprule
\multirow{2}{*}{\textbf{Methods}} & \multirow{2}{*}{\textbf{Backbone}}& \multicolumn{2}{c}{\textbf{Total}} \\ 
\cmidrule(lr){3-4}
& & \textbf{ROUGE-L(\%)} & \textbf{ROUGE-N(\%)}  \\ 
\midrule
\textbf{AstroSage-Llama-3.1-8B \cite{de2025achieving}} & MPT & 24.54 & 36.47 \\
\textbf{Mistral-7B \cite{samo2024fine}} & ViT & 26.93 & 39.64 \\
\midrule
\textbf{AstroSage-Llama-3.1-8B \cite{de2025achieving}+AstroRAG (Ours)}  & MPT & \textbf{\textcolor{blue}{28.51}} & \textbf{\textcolor{blue}{42.00}} \\
\textbf{Mistral-7B \cite{samo2024fine}+AstroRAG (Ours)}  & ViT & \textbf{\textcolor{red}{30.17}} & \textbf{\textcolor{red}{42.12}} \\
\bottomrule
\end{tabular}
}
\label{model_performance_table_ours}
\end{table}
\subsection{Quantitative Results}
Table \ref{model_performance_table} presents a comparative analysis of zero-shot question-answering performance on the AstroQA dataset across various large language models (LLMs) and their corresponding RAG enhancements. The baseline LLMs—Llama 2-Chat, Mistral-7B, and AstroSage-Llama-3.1-8B—achieve total accuracy scores of 28.06\%, 21.74\%, and 41.71\%, respectively, with corresponding F1-scores of 23.10, 9.20, and 41.45. Upon integrating the proposed AstroRAG, all models experience significant performance improvements. Notably, the Mistral-7B+AstroRAG variant achieves the highest total accuracy and F1-score of 79.49\%, outperforming all others across Easy (77.50\%, 77.45\%), Medium (80.28\%, 80.27\%), and Hard (83.59\%, 83.50\%) subsets. AstroSage-Llama-3.1-8B+AstroRAG also shows strong performance, with a total accuracy of 77.49\% and F1-score of 77.59\%. These results demonstrate that the proposed AstroRAG substantially enhances the reasoning capabilities of LLMs for astronomy, especially under more difficult question categories.
\begin{table}[tbp]
\centering
\small
\caption{Ablation study on the impact of max new tokens settings for the Mistral-7B + AstroRAG model, comparing accuracy (\%), F@1, and inference time (seconds). The results illustrate the trade-off between computational efficiency and performance, with smaller token budgets reducing latency but slightly affecting accuracy.}
\resizebox{\linewidth}{!}{
\begin{tabular}{c c c c c c c c c}
\toprule
\multirow{2}{*}{\textbf{Methods}} & \multirow{2}{*}{\textbf{\#Tokens}} & \multirow{2}{*}{\textbf{\#Time (s)}} & \multirow{2}{*}{\textbf{Acc (\%)}} &\multirow{2}{*}{\textbf{F@1}}\\ \\ 
\midrule
\multirow{5}{*}{\textbf{AstroRAG (Ours)}} & 16 &  0.47 & 78.18 & 78.23\\
  & 32 & 0.56 & 79.49 & 79.49\\ 
 &  64 &0.67 & 79.26 & 79.27\\
 & 128 & 0.72 & 79.11 & 79.12\\
  & 256 &0.72 & 78.95 & 78.97\\
\bottomrule
\end{tabular}
}
\label{ablation_table_inference_time}
\end{table} 

Table \ref{model_performance_table_sota} presents a detailed comparison of various RAG-based question answering methods evaluated on the AstroQA dataset, with results reported for Easy, Medium, Hard, and Total question subsets. The baseline method using Mistral-7B achieves the lowest performance, with an overall accuracy of 21.74\% and F1-score of 9.20\%, and similarly low scores across difficulty levels—20.74\%/8.60\% (Acc/F1) for Easy, 21.97\%/9.57\% for Medium, and 24.05\%/10.26\% for Hard. DrQA significantly improves performance, achieving 62.06\% accuracy and 62.43\% F1 on Easy questions, and 61.30\% accuracy and 61.55\% F1 overall. The Map-Reduce strategy shows further gains, particularly on Easy questions with 64.56\% accuracy and 64.94\% F1, while yielding 63.07\% accuracy and 63.48\% F1 overall. Notably, the proposed AstroRAG method surpasses all other approaches, achieving the highest scores across all categories: 77.50\%/77.45\% for Easy, 80.28\%/80.27\% for Medium, 83.59\%/83.50\% for Hard, and an overall performance of 79.49\% accuracy and 79.49\% F1. These results highlight the effectiveness of AstroRAG in leveraging both retrieval-augmented generation and pagerank-based re-ranking to significantly improve question answering performance on science-related queries.

\begin{figure}[tbp]
  \centering
  \includegraphics[width = \linewidth]{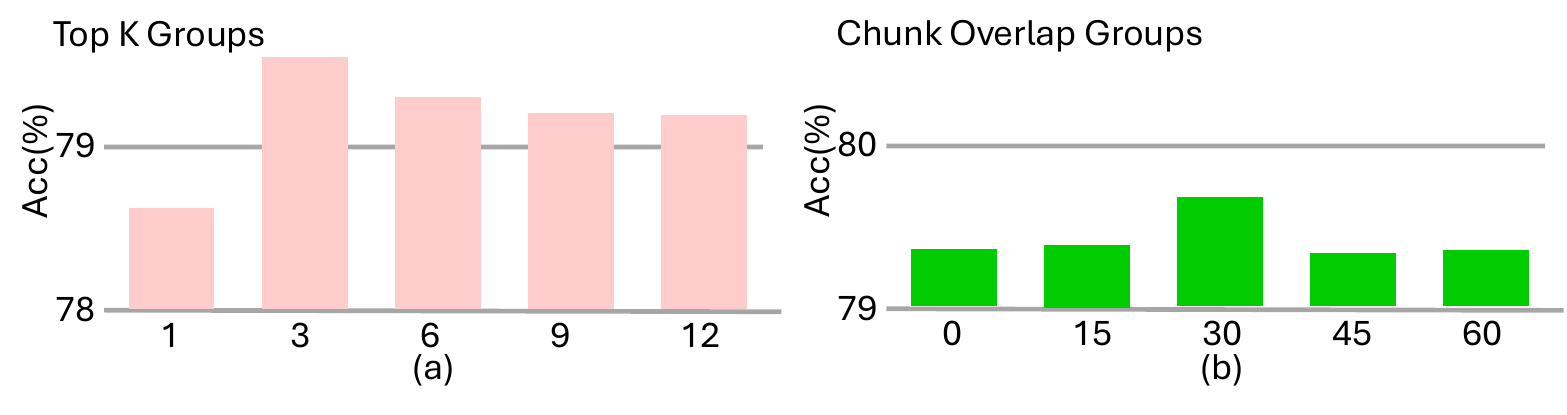}
  \vspace{-2em}
  \caption{Ablation study of key hyperparameters. (a) Effect of the number of pagerank top-$K$ groups ($K\in\{1,3,6,9,12\}$). (b) Effect of chunk-overlap groups ($\{0,15,30,45,60\}$). The best performance is observed at $K=3$ and an overlap of $30$.}
  \label{ablation_hyperparameters_top_k}
\end{figure}
Table \ref{model_performance_table_ours} presents a comparative analysis of zero-shot question-answering performance on the AstroRef-QA dataset across various LLM configurations using ROUGE-L and ROUGE-N evaluation metrics. The baseline AstroSage-Llama-3.1-8B model, built on the MPT backbone, achieves 24.54\% ROUGE-L and 36.47\% ROUGE-N, while the Mistral-7B baseline using the ViT backbone records slightly higher scores of 26.93\% and 39.64\% for ROUGE-L and ROUGE-N, respectively. Upon enhancement with the AstroRAG framework, both models exhibit significant performance gains. Specifically, AstroSage-Llama-3.1-8B + AstroRAG improves to 28.51\% ROUGE-L and 42.00\% ROUGE-N, and Mistral-7B + AstroRAG further achieves the highest scores of 30.17\% ROUGE-L and 42.12\% ROUGE-N. These results demonstrate the effectiveness of the AstroRAG integration in improving both lexical overlap and content relevance in generated answers.
\subsection{Ablation Study}
\begin{figure*}[tbp]
  \centering
  \includegraphics[width = \linewidth]{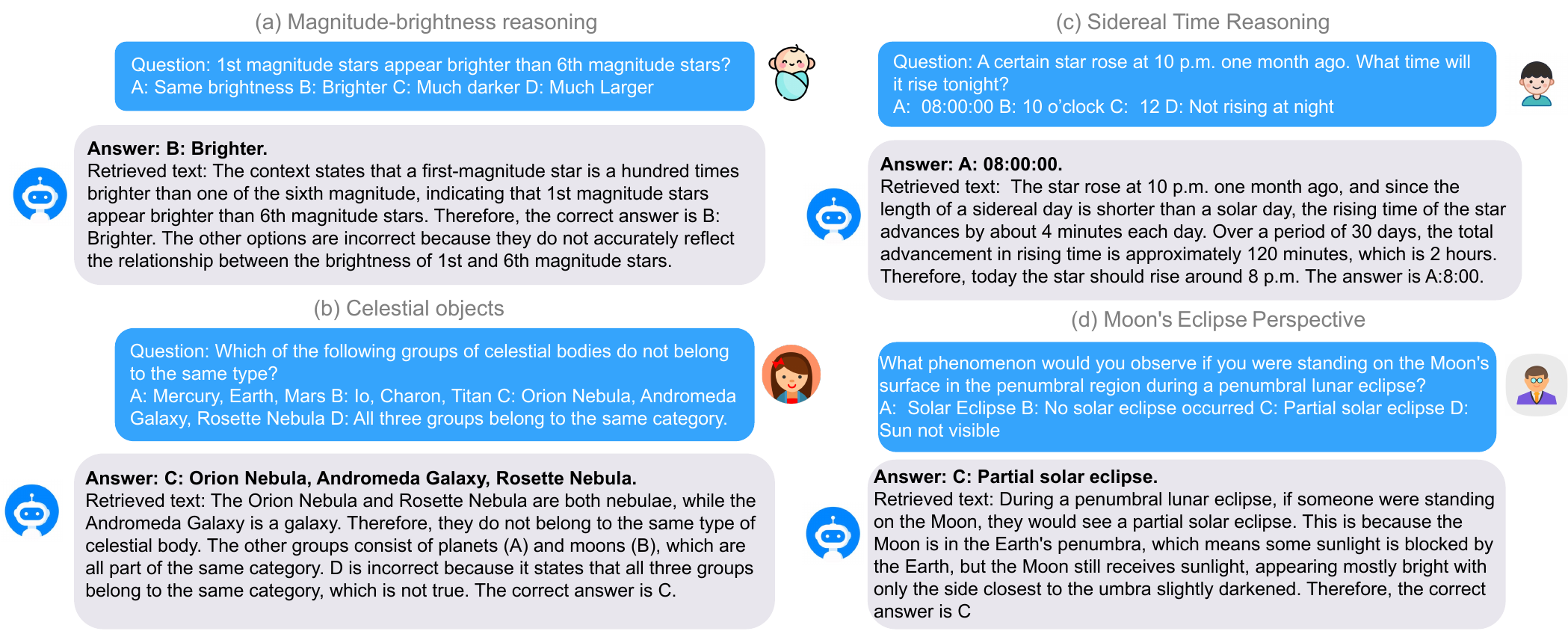}
  \caption{\textbf{Qualitative observations on retrieved-context question answering.} Four representative queries from astronomy and geography are shown (blue). For each query, the model selects an answer (gray) and provides supporting retrieved passages, illustrating: (a) magnitude–brightness reasoning, (b) category discrimination among celestial objects, (c) sidereal time reasoning, and (d) time-zone conversion.
}
  \label{ablation_visual_examples}
\end{figure*}

Table \ref{ablation_table_inference_time} presents an ablation study assessing the impact of different max new tokens settings on the performance of the Mistral-7B + AstroRAG model. The results demonstrate a trade-off between computational latency and prediction performance, with varying token budgets. As the number of output tokens increases from 16 to 64, both accuracy and F1-score improve significantly—from 78.18\% to 79.26\% in accuracy and from 78.23 to 79.27 in F1—while the inference time grows moderately from 0.47 to 0.67 seconds. Further increases in token count to 128 and 256 tokens lead to minimal gains or even slight declines: accuracy reaches 79.11\% and 78.95\%, respectively, while F1 remains stable at 79.12 and drops marginally to 78.97. The inference time also plateaus at 0.72 seconds for 128 and 256 tokens. These results indicate that setting the token limit to 64 strikes an optimal balance between latency and model performance, maximizing accuracy and F1 without incurring significant computational cost.

\subsection{Impacts of key hyperparameters}
To understand the impact of key hyperparameters on model performance, Fig.~\ref{ablation_hyperparameters_top_k} presents an ablation study evaluating two crucial settings: the number of pagerank top-$K$ groups and the degree of chunk-overlap. In subfigure~(a), varying the top-$K$ groups across values $K \in \{1, 3, 6, 9, 12\}$ reveals a clear trend: the highest accuracy (approximately $79.5\%$) is achieved when $K = 3$, with performance consistently dropping as $K$ increases, reaching below $79.1\%$ for $K = 12$. Subfigure~(b) investigates the chunk-overlap parameter with values $\{0, 15, 30, 45, 60\}$, showing that an overlap of $30$ yields the best accuracy (around $79.5\%$). In contrast, both lower ($0$) and higher ($45$, $60$) overlap values result in slightly reduced accuracy (around $79.2\%$). These results highlight that small pagerank top k and moderate chunk overlap are optimal for maximizing accuracy.

\subsection{Qualitative Observations}
Fig.~\ref{ablation_visual_examples} presents qualitative case studies of our retrieved-context question answering system. Each subpanel displays a natural-language multiple-choice query (blue), the model’s selected answer with brief justification (gray), and the supporting passages returned by retrieval. The examples span (a) magnitude–brightness reasoning in astronomy, (b) category discrimination among celestial objects, (c) temporal reasoning, and (d) moon's eclipse perspective. Together, these cases illustrate how retrieval grounds the model’s predictions, enabling domain knowledge recall and step-wise explanation while highlighting the breadth of reasoning skills exercised across physics, astronomy, and geography.

\section{Conclusion}
We introduced AstroRAG, a training-free, fully local RAG pipeline that couples token-aware chunking with two-stage retrieval—MMR followed by reader-driven PageRank—to form compact, mutually supportive contexts under a strict token budget. By indexing per instance in an ephemeral Elasticsearch store and logging both predictions and evidence, the system is privacy-preserving, reproducible, and resistant to cross-task leakage. On AstroQA, the RAG-enhanced Mistral-7B attains 79.49\% accuracy and 79.49\% F1, nearly doubling its non-RAG baseline and underscoring the impact of disciplined retrieval on domain reasoning. Remaining limitations include sensitivity to embedding model choice and retrieval hyperparameters. Future work will explore adaptive chunking and $k$ selection, calibrated or learned re-ranking, multi-hop aggregation across documents, OCR for scanned PDFs, multilingual inputs, and extensions to open-ended question answering.

\section{Acknowledgment}
We would like to thank A. Prof. Kee Siong Ng at the Australian National University and Dr. Ioana Ciuca at Stanford University for their valuable comments and suggestions. This work is part of the collaboration between Learning Machines Pty Ltd and the Advanced Instrumentation Technology Centre at the Australian National University.

\bibliographystyle{IEEEtran} 
\bibliography{rag_llms}

\end{document}